\documentclass[10pt,twocolumn,letterpaper]{article}

\usepackage[pagenumbers]{cvpr} % To force page numbers, e.g. for an arXiv version

\usepackage{graphicx}
\usepackage{amsmath}
\usepackage{amssymb}
\usepackage{booktabs}
\usepackage{float}

\usepackage[pagebackref,breaklinks,colorlinks]{hyperref}

\usepackage[capitalize]{cleveref}
\crefname{section}{Sec.}{Secs.}
\Crefname{section}{Section}{Sections}
\Crefname{table}{Table}{Tables}
\crefname{table}{Tab.}{Tabs.}

\def\cvprPaperID{2290} % *** Enter the CVPR Paper ID here
\def\confName{CVPR}
\def\confYear{2022}

\title{FaceVerse: a Fine-grained and Detail-controllable 3D Face Morphable Model \\ from a Hybrid Dataset}

\author{
Lizhen Wang\textsuperscript{1}, Zhiyuan Chen\textsuperscript{2}, Tao Yu\textsuperscript{1}, Chenguang Ma\textsuperscript{2}, Liang Li\textsuperscript{2}, Yebin Liu\textsuperscript{1}\\
Tsinghua University\textsuperscript{1}, Beijing, China\\
Ant Group\textsuperscript{2}, Hangzhou, China
}
\begin{document}

\twocolumn[{%
\maketitle
\begin{center}
\includegraphics[width=0.9\linewidth]{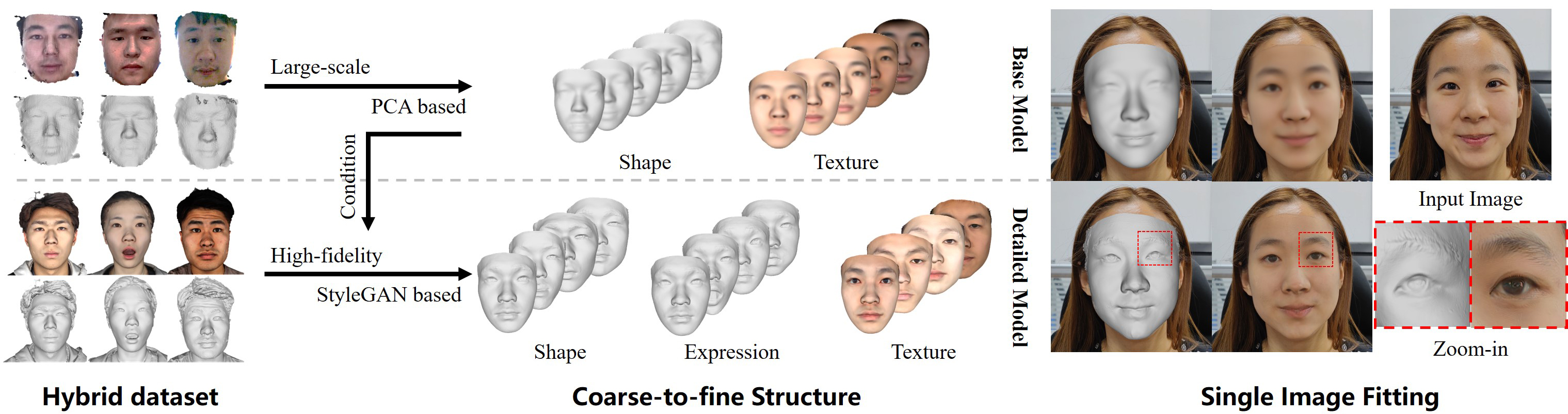}
\captionof{figure}{Our hybrid dataset, the base and detail model of FaceVerse, as well as our single-image fitting result.}
\label{fig:maindescrip}
\end{center}%
}]

%%%%%%%%% ABSTRACT
\begin{abstract}
We present FaceVerse, a fine-grained 3D Neural Face Model, which is built from hybrid East Asian face datasets containing 60K fused RGB-D images and 2K high-fidelity 3D head scan models. A novel coarse-to-fine structure is proposed to take better advantage of our hybrid dataset. In the coarse module, we generate a base parametric model from large-scale RGB-D images, which is able to predict accurate rough 3D face models in different genders, ages, etc. Then in the fine module, a conditional StyleGAN architecture trained with high-fidelity scan models is introduced to enrich elaborate facial geometric and texture details. Note that different from previous methods, our base and detailed modules are both changeable, which enables an innovative application of adjusting both the basic attributes and the facial details of 3D face models. Furthermore, we propose a single-image fitting framework based on differentiable rendering. Rich experiments show that our method outperforms the state-of-the-art methods.
 \end{abstract}

%%%%%%%%% BODY TEXT 
\section{Introduction}
\label{sec:intro}
3D human face modeling has been a hot topic in computer vision and computer graphics, which enables a wide range of applications such as film, video games, mixed reality, etc. Since 3D Morphable Model (3DMM)~\cite{CGIT1999Blanz} was proposed in 1999, it has been one of the most powerful tools in face-related researches due to its effective control of facial shape, expression and texture. 
However, recent researches pose more challenges to 3DMM in terms of accuracy, photo-realistic details and editability. On one hand, the performance of 3DMMs is limited due to the difficulty of data acquisition. On the other hand, given a coarse face model, detailed facial geometry and texture are still not changeable in the previous methods~\cite{CVPR2020Facescape, CVPR2020LFPF}, which limits the detailed adjustment of facial features. 
To overcome the above issues, we propose a hybrid dataset and design a coarse-to-fine structure to combine high generalization ability and fidelity. Furthermore, facial geometry and texture details, like small changes of facial features, can also be parameter-changeable.

At one end of the spectrum, existing 3D face datasets are usually limited in either scale or fidelity. The capturing system can be divided into two categories: sparse or dense camera arrays~\cite{CVPR2020Facescape, CVPR2016LSFM, ICCV2017LYHM, CVPR2005FRGC, CVPR2020LFPF} and consumer depth sensors~\cite{TVCG2014FaceWarehouse, AVSBS2009BFM, ICB2016, TIP2021FromX}. The former system requires elaborated setup and the data collection process is quite time-consuming, which limits the scale of captured dataset to a few hundreds. The latter system is off-the-shelf and takes less time in data acquisition, which allows collecting RGB-D data from a large number of identities. However, the captured RGB-D data usually suffers from low resolution and low precision. The insufficiency of scale or fidelity limits the performance of previous works in either generalization or fidelity. Therefore, we propose to build a hybrid dataset.

At the other end, the formulation of previous 3DMMs can not represent parameter-changeable facial details. PCA-based methods~\cite{AVSBS2009BFM, CVPR2016LSFM, li2017learning, ICB2016, TIP2021FromX} can describe shape and expression changes in an effective way. Multi-linear methods~\cite{CVPR2005FRGC, TVCG2014FaceWarehouse} present a larger parameter space to cover more information of the corresponding datasets. Non-linear methods~\cite{CVPR2018nonlinear, jiang2019disentangled} use neural networks to achieve better flexibility. However, all the above methods can not represent the facial details, like the detailed shape of facial features. Recent methods~\cite{CVPR2020Facescape, CVPR2020LFPF} show strong capability in 3D fine-grained face model reconstruction, but they still rely on pre-trained super-resolution or displacement prediction networks, which means the facial details are not parameter changeable. To conclude, a 3DMM representation with changeable facial details has not been proposed yet.

To overcome the limitations above, in this paper, we propose FaceVerse, which achieves high generalization ability and fidelity using a hybrid dataset and can generate parameter-changeable facial details. Firstly, we collect a hybrid dataset of East Asians consisting of a large-scale dataset captured by consumer depth sensors and a high-fidelity dataset captured by a multi-camera system. Secondly, we propose a coarse-to-fine structure to scheme our parametric model. The base model is first built from the large-scale dataset, which guarantees strong generalization ability and basic fidelity of the base model. Then, input with the UV maps unwrapped from the base model, we build our detailed model using a novel conditional StyleGAN architecture, which can generate changeable facial details input with additional latent code and noise while preserving the basic facial attributes provided by the input base model. Different from the original StyleGAN~\cite{karras2020a, karras2020analyzing}, our generator takes advantage of multi-scale features encoded from the input maps to constrain the output maps and we use an additional normal discriminator to further enrich the geometry details. Note that two conditional StyleGAN networks are used in two phases: detail generation and expression refinement. Finally, we propose a single-image fitting pipeline based on differentiable rendering, which also follows the coarse-to-fine idea. Benefiting from the hybrid dataset, the coarse-to-fine scheme and the novel conditional StyleGAN architecture, the proposed FaceVerse shows better performance than previous 3DMM methods both qualitatively and quantitatively.

Our contributions are summarized as follows:
\begin{itemize}
\item We build a hybrid dataset and propose a coarse-to-fine scheme to make better use of the dataset: the large-scale RGB-D dataset guarantees high generalization ability of our base model and the high-fidelity scan dataset helps to enrich the geometry and texture details of our detailed model.
\item We propose a conditional StyleGAN architecture with normal discriminators, which allows changing facial details while preserving basic facial attributes.
%\item A monocular 3D face fitting pipeline is proposed to recover high-fidelity and fine-grained 3D face model from a single image. 
\item The proposed FaceVerse provides a powerful tool for face modeling of East Asians and we have released our pre-trained models and the detailed dataset to public for research purpose\footnote{https://github.com/LizhenWangT/FaceVerse}.
\end{itemize}

\section{Related Work}
\label{sec:related work}

\paragraph{3D Face Morphable Model.} 
The 3D face morphable model (3DMM) has been a long-standing research topic in computer vision since first proposed by Blanz et al.~\cite{CGIT1999Blanz} in 1999. 3DMM was first formulated as a linear model by the PCA algorithm, which can represent the shape and texture of 3D face model. The following researches~\cite{AVSBS2009BFM, IJCV2018lsfm, TVCG2014FaceWarehouse, TOG2017flame, CVPR2020Facescape, bao2021tog, CVPR2020LFPF} improved the performance using larger 3D face datasets. Moreover, new representations including multi-linear and non-linear models for 3DMM were also proposed in ~\cite{SIGGRAPH2006multilinear, TOG2013sparse, ECCV2014multilinear, Tog2010example, CVPR2018self, CVPR2019towards, CVPR2018nonlinear, CVPR2020LFPF}.

Recent 3D face datasets show higher diversity in both identities and expressions. LSFM~\cite{IJCV2018lsfm} was built from a large 3D face dataset containing 10,000 face scans and shows better generalization in facial shape fitting. In the meanwhile, 3D face datasets with rich expressions were also collected to incorporate the facial expression bases into 3DMM~\cite{SIGGRAPH2006multilinear, TVCG2014FaceWarehouse, TOG2017flame, CVPR2020Facescape, bao2021tog}. Furthermore, with the development of elaborated capturing system like dense camera arrays, recent 3DMM methods\cite{CVPR2020LFPF, CVPR2020Facescape, bao2021tog} exhibited even higher accuracy in 3D face modeling.

Besides the improvement in 3D face datasets, novel modeling mechanisms were also presented for better performance and flexibility. Vlasic et al.~\cite{SIGGRAPH2006multilinear} first proposed a multi-linear model to jointly estimate the variations in identity and expression,  Cao et al.~\cite{TVCG2014FaceWarehouse} and Yang et al.~\cite{CVPR2020Facescape} built comprehensive bilinear models which decompose the face meshes in both identity and expression dimensions. Recently, non-linear models were also proposed to enable adaptive and high-level facial deformations. Neumann et al.~\cite{TOG2013sparse} decomposed the captured face mesh sequences into the sparse and localized deformation components. With the development of neural networks, generative adversarial networks (GAN) were also used to build the non-linear 3DMMs~\cite{CVPR2018VAE, CVPR2019towards, CVPR2020LFPF, GALTERI201931}, the face representations of which can be controlled by high-level semantics.
% 

% 自从3DMM被提出以来
\paragraph{Monocular Face Reconstruction Based on 3DMM.}

Monocular 3D face reconstruction based on 3DMM plays an important role in many applications like face alignment~\cite{eccv2018prn, cvpr2016face, eccv2020towards} and face view synthesis~\cite{CVPR2016deepstereo, CVPR2018view}. With the assistance of 3DMM, the 3D face reconstruction task can be simplified as a model fitting problem. Early methods~\cite{CVPR2005Estimating, AVSBS2009BFM, CVPR2016face2face} mainly tried to regress the parameters of 3DMM using the facial landmarks or some other facial features. Then the convolutional neural networks were used to directly predict the parameters from an input face image~\cite{CVPR2017end2end, cvpr2016face, CVPR2017regressing, eccv2018prn, tewari2017mofa, eccv2020towards}. Recently, self-supervised methods~\cite{deng2019accurate, chen2020self} based on differentiable rendering were presented and show great performance in fitting 3D face models from a single face image.

The above methods based on model parameters prediction are limited in representing facial details, and thus multi-layer refinement structures are proposed to reconstruct detailed face models. Recent works~\cite{richardson2017learning, sela2017unrestricted, chen2019photo, huynh2018mesoscopic, tran2018extreme, CVPR2020Facescape, TOG2021deca} firstly generated a rough face model through the model parameters prediction, and then refined the facial details by adjusting the rendered depth or predicting a displacement map. Lin et al.~\cite{bao2021tog} generated high-fidelity models by the optimization of the albedo and normal maps. However, the detailed facial features are still not parameter-changeable in these researches, which limits the adjustment of facial details in 3D face models.

Compared with the state-of-the-art 3DMM and monocular facial reconstruction methods, our approach is superior in the following aspects: (a) our model is built from a hybrid dataset, which contains a large-scale coarse dataset and a high-fidelity detailed dataset; (b) we propose a coarse-to-fine model which consists of a PCA-based model and a novel conditional styleGAN-based non-linear model; (c) our coarse-fine model-fitting pipeline based on differentiable rendering can not only reconstruct high-fidelity 3D face models from in-the-wild face images, but also generate facial details which can be adjusted by our detailed parameters.

\section{Hybrid Dataset}
\label{sec:dataset}
\subsection{Coarse Dataset}
\begin{figure}
    \centering
  	\includegraphics[width=0.98\linewidth]{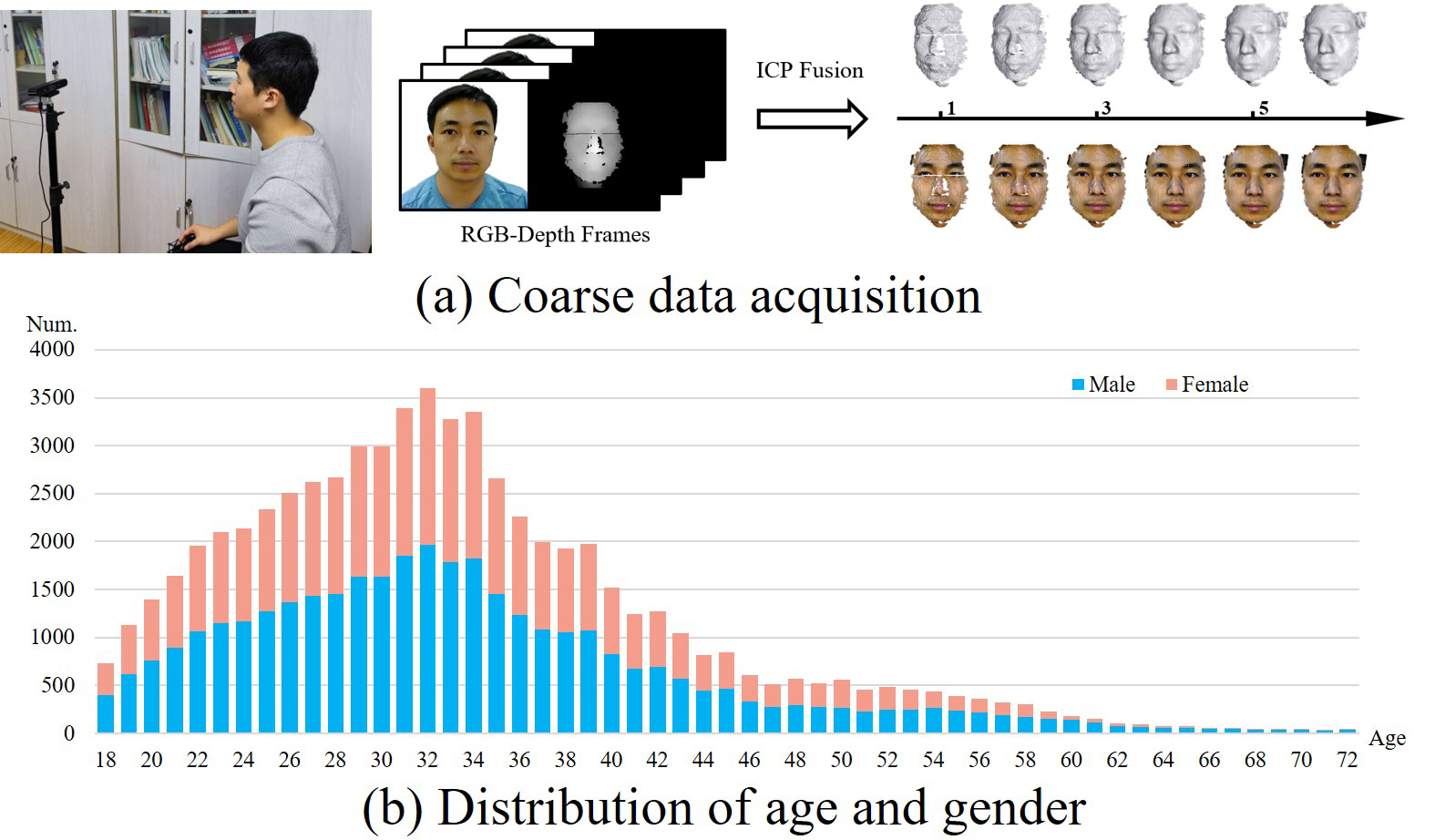}
    	\captionof{figure}{The coarse data acquisition process and the age and gender distribution of our coarse dataset.}
	 \label{fig:rawcoarse}
\end{figure}

We chose the structured-light depth sensors to collect coarse 3D face data from volunteers, which show better performance than ToF-based devices in distance below 1 meter. Compared with dense camera array, the structured-light depth sensors are cost-friendly and more convenient for parallel setup, which allows collecting RBG-D data from a large number of identities. In practice, as shown in Fig.~\ref{fig:rawcoarse}.a, we collect about 5 RGB-D frames for each volunteer and the frames are fused by ICP registration to generate a smooth facial point cloud. The whole capturing process for each volunteer only costs 5 to 10 seconds. With the assistance of several data acquisition companies and parallel capturing, we finally get 60K textured facial point clouds of East Asians after data cleaning. Volunteers are required to keep neutral expression during the capturing to ensure the consistence of data distribution in expression. 

In order to generate a topologically uniformed parametric model, we use a pre-designed 3D facial template mesh to fit the point clouds. We firstly detect facial landmarks using OpenSeeFace\footnote{https://github.com/emilianavt/OpenSeeFace} from the captured RGB images and project them to the fused point clouds. Then we roughly align the point clouds to our template mesh by 3D landmarks. Finally, a Non-rigid ICP algorithm~\cite{li2008global} is utilized to deform the template mesh to the aligned point clouds. The distribution of age and gender is presented in Fig.~\ref{fig:rawcoarse}.b.

\subsection{Detailed Dataset} 
\begin{figure}
    \centering
  	\includegraphics[width=0.95\linewidth]{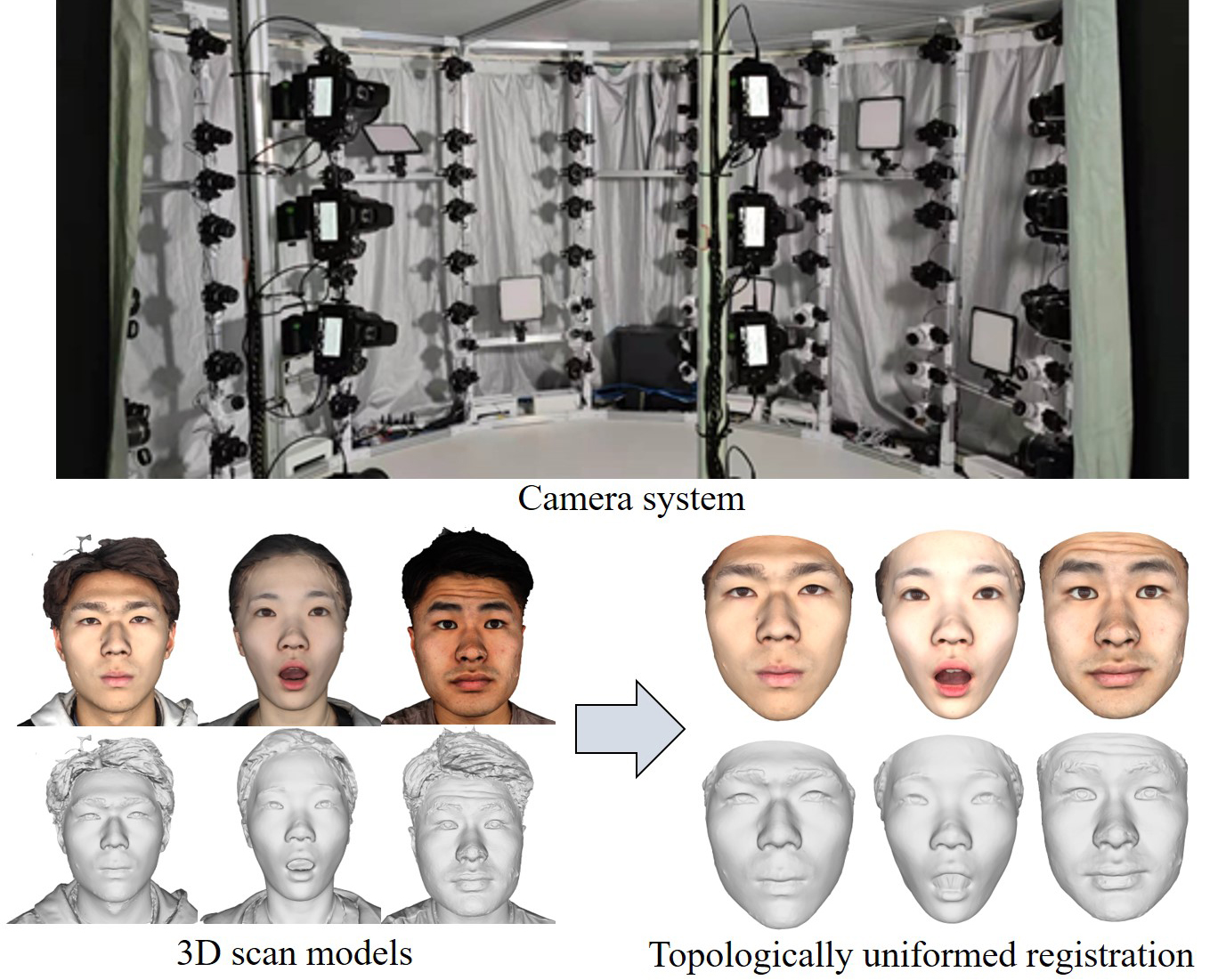}
    	\captionof{figure}{Our camera system for data collection, as well as detailed 3D scan models and corresponding registration results.}
	 \label{fig:detailenvoriment}
\end{figure}

Our camera system for 3D scan model collection consists of 128 DSLR cameras, which equip with 85 mm lenses and are placed about 2.5 meters away from the volunteer, as shown in Fig.~\ref{fig:detailenvoriment}. The cameras are arranged in cylinder facing towards the center by 16 pillars with 8 cameras on each, which is similar to high quality full body scan system in~\cite{zheng2021deepmulticap, yu2021function4d}. During data collection, 128 images with $6000\times4000$ resolution will be synchronously collected from different view points. We follow the Data acquisition process of FaceWarehouse~\cite{TVCG2014FaceWarehouse}, where the volunteers are required to perform 21 specific expressions including neutral expression. We finally collect 2,310 scan models (110 identities in 21 expressions) for training and 378 scan models (18 identities in 21 expressions) for testing, which has been released to public for research purpose.

After the data collection, the 3D scans are fitted to our topologically uniformed template. Firstly, 3D landmarks are marked for rigid-ICP alignment by projecting 2D landmarks onto the 3D scans. Our base model generated from the coarse dataset (Sec.~\ref{sec:basemodel}) is used to fit the scans with corresponding 3D landmarks. Then, the resulting fitted models are up-sampled in the UV space (from $200\times200$ to $1024\times1024$) for the subsequent registration. Finally, we conduct the detailed deformation on the fitted models using Non-rigid ICP ~\cite{li2008global}.

\section{FaceVerse Model}
\label{sec:model}

A coarse-to-fine scheme is proposed to generate the proposed model, FaceVerse, from the hybrid dataset: our base model is built from the large-scale coarse dataset by PCA and the detailed model is built from the high-fidelity detailed dataset by our conditional StyleGAN networks. In addition, we also present a single-image fitting framework based on differentiable rendering.
  
\subsection{Base Model Generation}
\label{sec:basemodel}

We use the classical data dimension reduction algorithm, PCA, to build the shape and texture models from the large-scale coarse dataset, which guarantees the high generalization ability and basic fidelity. The first 100 shape principal components and the first 200 texture principal components are preserved in our base model. Note that, in order to improve our performance on cheeks, which are almost invisible in the coarse RGB-D frames, we add the first 20 shape principal components learned from the detailed dataset into the base shape model. As a result, our base model can be expressed by shape parameters $p_{shape}=\{s_1, s_2, ... , s_m\}\in\mathbb{R}^{m}$ and texture parameters $p_{tex}=\{t_1, t_2, ... , t_k\}\in\mathbb{R}^{k}$:
\begin{small}
\begin{equation}
  S_{base} = \overline{S}  + \sum\limits_{i = 1}^{m} s_{i} \alpha_{i}  \quad T_{base}= \overline{T}  + \sum\limits_{i = 1}^{k} t_{i} \beta_{i}
  \label{eq:pca}
\end{equation}
\end{small}
where $m=120$, $k=200$ and $\overline{S}$ \& $\overline{T}$ denotes mean shape and texture. The shape and texture principal components are represented by the shape vectors $\{\alpha_{1}, \alpha_{2}, ... , \alpha_{m}\}$ and the texture vectors $\{\beta_{1}, \beta_{2}, ... , \beta_{k}\}$, where $\alpha_{i}\in\mathbb{R}^{3n}$ and $\beta_{i}\in\mathbb{R}^{3n}$ ($n$ denotes the number of vertices).

As the coarse faces are captured in neutral expressions, the expression model is generated from the detailed dataset using PCA. The fisrt 64 principal components are used in our expression base model, which can be formulated by expression parameters $p_{exp}=\{e_1, e_2, ... , e_l\}\in\mathbb{R}^{l}$ and expression vectors $\{\gamma_{1}, \gamma_{2}, ... , \gamma_{l}\}$, where $l=64$ and $\gamma_{i}\in\mathbb{R}^{3n}$. As a result, our base model can be formulated as
\begin{small}
\begin{equation}
\begin{aligned}
  M_{base} = \{S,T| & S=\overline{S} + \sum\limits_{i = 1}^{m} s_{i} \alpha_{i} + \sum\limits_{i = 1}^{l} e_{i} \gamma_{i},\\
                    & T=\overline{T}  + \sum\limits_{i = 1}^{k} t_{i} \beta_{i}\} 
\label{eq:base}
\end{aligned}
\end{equation}
\end{small}

Benefiting from the large-scale coarse dataset, our base model shows strong performance in fitting faces of different ages and genders quantitatively. However, our base model can not preserve the facial geometry and texture details, which will be generated by the following detailed model.

\subsection{Detailed Model Generation}
\label{sec:finemodel}

\begin{figure*}[ht]
    \centering
  	\includegraphics[width=0.99\linewidth]{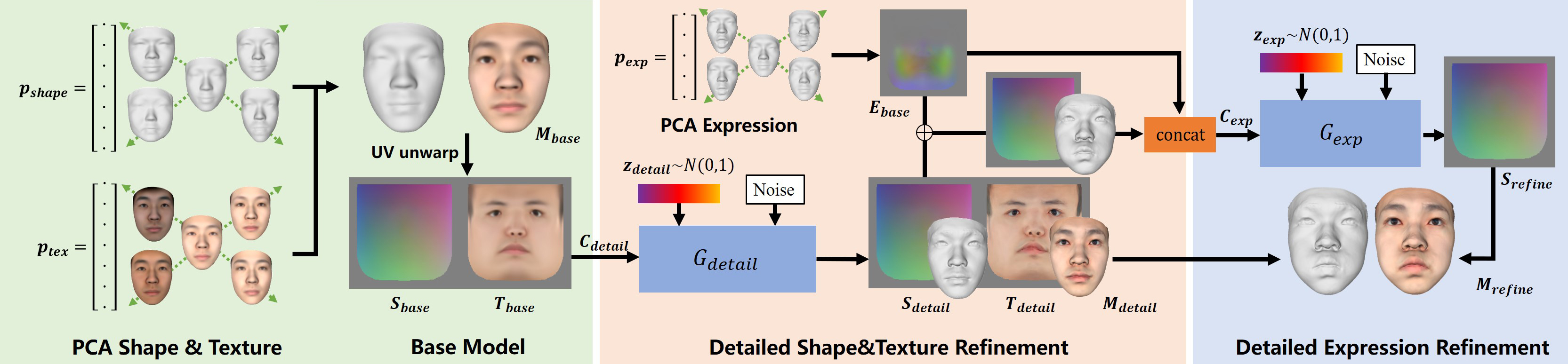}
    	\captionof{figure}{FaceVerse model generation pipeline. Using the base PCA model, we first unwrap the base model $M_{base}$ into UV maps. Then the detail generator $G_{detail}$ helps to enrich facial details controlled by the additional latent code $z_{detail}$ and the injected noise. Finally, the expression-related geometry changes will be further refined by another expression refinement generator $G_{exp}$ input with the additional latent code $z_{exp}$ and the injected noise.}
	 \label{fig:pipeline}
\end{figure*}

As shown in Fig.~\ref{fig:pipeline}, to incorporate more detailed facial geometry and texture, we propose a neural representation for our detailed model, which can take better advantage of the detailed dataset. The base model is first unwrapped into the UV space and up-sampled to $1024 \times 1024$ to facilitate subsequent processing. The whole refinement work is divided into a shape\&texture refinement part and a expression refinement part.

\begin{figure}
    \centering
  	\includegraphics[width=0.98\linewidth]{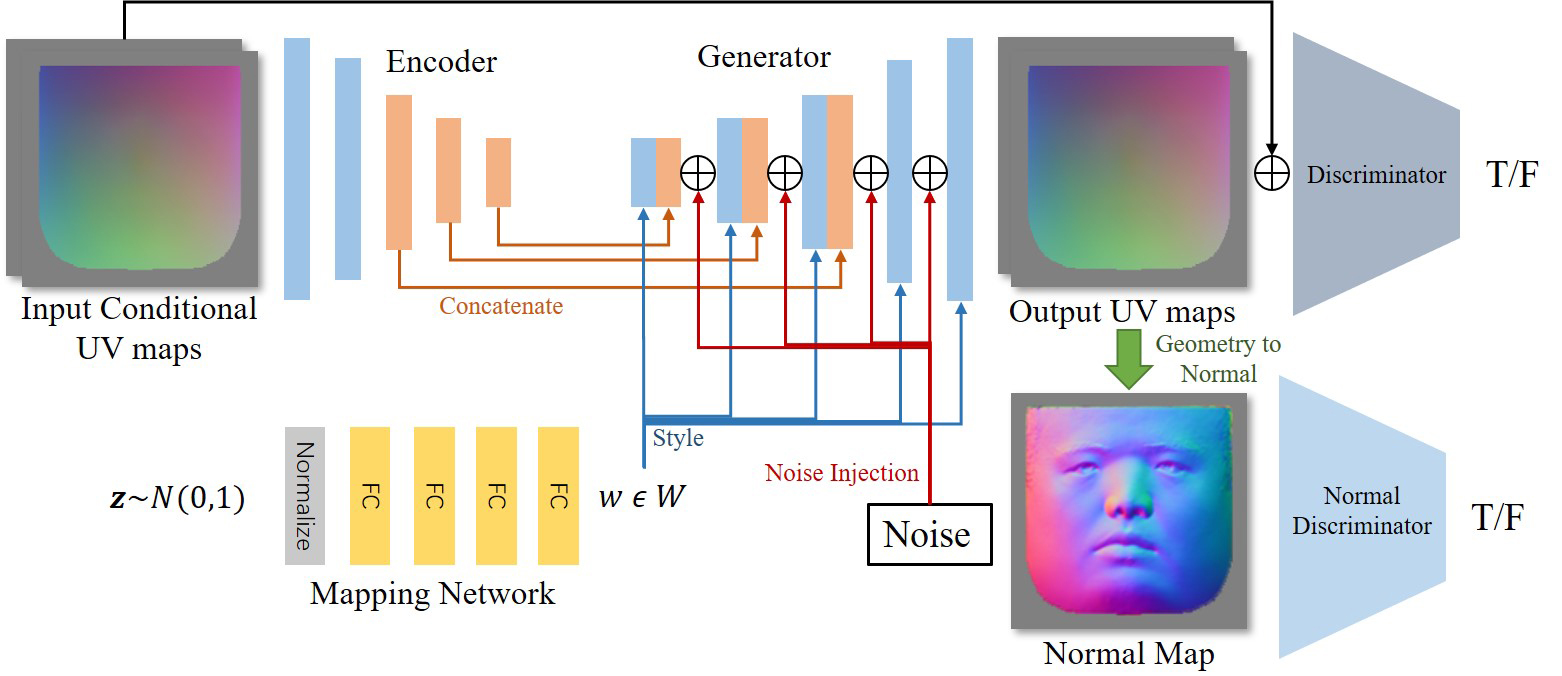}
    	\captionof{figure}{The architecture of our conditional StyleGAN network.}
	 \label{fig:uvgan}
\end{figure}

To better generate the facial details while preserving the basic facial attributes provided by our base model, we propose a conditional StyleGAN network. As shown in Fig.~\ref{fig:uvgan}, we adopt the generator, mapping network and noise injection module of StyleGAN and design an extra encoder to encode multi-scale features from the input UV maps. The multi-scale features are added into the generator as a conditional input, which helps to constrain the similarity of the input and output UV maps. Besides, we use two discriminators in our conditional StyleGAN: one discriminator input with the input and output UV maps, which helps to generate more details and constrain the similarity of input and output maps; another normal discriminator input with a UV normal map calculated from the output geometry, which helps to generate more geometry details and constrain a reasonable neighborhood relationship of the adjacent points. The input detail latent code $z\in\mathbb{R}^{512}$ is sampled from the standard normal distribution, which will be disentangled into the style inputs by the mapping network. In the meanwhile, random noise is injected to enrich tiny details like beard and eyebrows.

In the shape\&texture refinement part, we use a conditional StyleGAN $G_{detail}$ to generate facial geometry and texture details. Firstly, the input base model $M_{base}$ in the neutral expression is unwrapped into a geometry UV map $S_{base}$ and a texture UV map $T_{base}$. Note that we believe geometry details and texture details should have a strong correlation, so we concatenate the geometry and texture UV maps into a 6-channel input $C_{detail}$. Due to the combined training of geometry and texture channels, the output geometry and texture are influenced by each other, which further facilitates the subsequent detailed geometry fitting (Sec.~\ref{sec:fitting}). The concatenated 12-channel input and output UV map of $G_{detail}$ is entered into the discriminator $D_{detail}$ and a 3-channel UV normal map is entered into the normal discriminator $Dn_{detail}$.
As discussed in Sec.~\ref{sec:ablation}, the output detailed model $M_{detail}$ shows fine-grained facial geometry and texture details, which can be controlled by $z_{detail}$ and the injected noise. Moreover, the basic shape and texture provided the input base model are still retained. The loss terms used in the training of $G_{detail}$ can be formulated as
\begin{small}
\begin{equation}
\begin{aligned}
 \mathcal{L}_{detail} = & \lambda_s \|S_{base}-S_{detail}\|^{2} + \\
&\lambda_t \|T_{base}-T_{detail}\|^{2} + \mathcal{L}_{GAN}
 \label{eq:training_loss} 
\end{aligned}
\end{equation}
\end{small}
where $\mathcal{L}_{GAN}$ represents the adversarial loss term and the path length regularization term of StyleGAN provided by $D_{detail}$ and $Dn_{detail}$. Note that our training process is under incomplete supervision and thus the used training data contains not only the data pairs from the detailed dataset but also the conditional UV maps generated from our coarse dataset, which further guarantees the effective interpolation capability of our detail generator $G_{detail}$.

In the expression refinement part, detailed expression-related geometry changes like a smiling mouth will be further refined by another conditional StyleGAN network $G_{exp}$. Given the detailed geometry UV map in neutral expression $S_{detail}$ and the base expression formulated by the UV offset map $E_{base}$, $G_{exp}$ will refine the detailed geometry while preserving the basic shape and expression. Specifically, the 6-channel conditional input $C_{exp}$ consists of a basic geometry which is the sum of $S_{detail}$ and $E_{base}$ and an additional expression offset $E_{base}$, where the basic geometry input is used to constrain the similarity of the input and output geometry and the additional expression input provides priors of the facial expression. The concatenated 9-channel input and output UV map of $G_{exp}$ is entered into the discriminator $D_{exp}$ and the concatenated 6-channel UV map consisting of a normal map and an expression offset map is entered into the normal discriminator $Dn_{exp}$.
After training of $G_{exp}$, the 3-channel output geometry $S_{refine}$ can represent more detailed expression changes, as discussed in Sec.~\ref{sec:ablation}, and the generation is also controlled by the latent code $z_{exp}$ and injected noise. The training process of $G_{exp}$ utilizes the paired data generated from our detailed dataset and the training loss terms can be formulated as
\begin{small}
\begin{equation}
\begin{aligned}
 \mathcal{L}_{exp} = & \lambda_e \|S_{detail}+E_{base}-S_{refine}\|^{2} + \mathcal{L}_{GAN}
 \label{eq:training_loss2} 
\end{aligned}
\end{equation}
\end{small}

\subsection{Coarse-to-Fine Single-Image Fitting}
\label{sec:fitting}

\begin{figure}
    \centering
  	\includegraphics[width=0.98\linewidth]{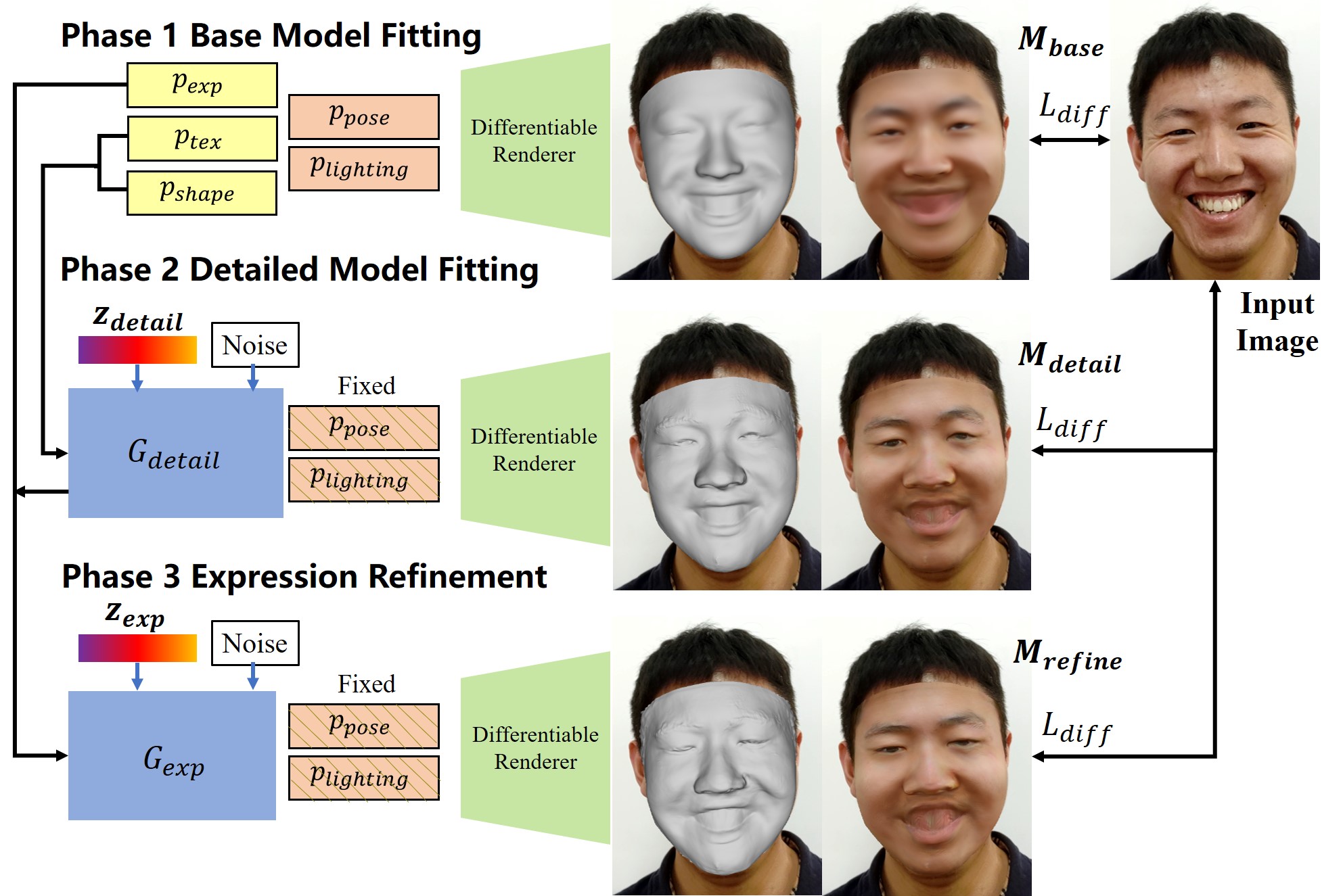}
    	\captionof{figure}{Our single-image fitting pipeline based on differentiable rendering.}
	 \label{fig:fit_pipeline}
\end{figure}

We further propose a single-image fitting pipeline, which adopts an optimization algorithm based on differentiable rendering, in this subsection, as shown in Fig.~\ref{fig:fit_pipeline}. The fitting process is divided into three phases: base model fitting, detailed model fitting and expression refinement.

In the first base model fitting phase, the parameters to be optimized include $p_{shape}$, $p_{tex}$, $p_{exp}$ of our base model and additional pose\&lighting parameters. The pose parameters $p_{pose}\in\mathbb{R}^{6}$ control the 3-dimensional translation and the 3-dimensional rotation, which are expressed in Euler angles. We use the first three bands of Spherical Harmonics(SH)~\cite{basri2003lambertian} for the definition of lighting parameters $p_{lighting}\in\mathbb{R}^{27}$. Our optimization loss terms can be formulated as
\begin{small}
\begin{equation}
\mathcal{L}_{diff} = \mathcal{L}_{lms} + \mathcal{L}_{photo} + \mathcal{L}_{reg}
 \label{eq:basefit} 
\end{equation}
\end{small}
where $\mathcal{L}_{lms}$ denotes the mean square loss of the detected 2D facial landmarks and the projected landmarks from the 3D model, $\mathcal{L}_{photo}$ denotes the mean square loss of the rendered image and the input image and $\mathcal{L}_{reg}$ denotes the L2 regular terms of $p_{shape}$, $p_{tex}$ and $p_{exp}$. The resulting shape, texture and expression are unwrapped into UV maps for subsequent phases.

In the detailed model fitting phase, we recover the identity-related facial geometry and texture details through the optimization of our pre-trained detail generator $G_{detail}$. The expression offset UV map, $p_{pose}$ and $p_{lighting}$ generated from the previous phase are fixed in this and the next phase. Input with the UV maps of shape and texture generated from the previous phase, our detail latent code $z_{detail}$ and the injected noise are first randomly sampled and then optimized by differentiable rendering with the similar loss terms, where $\mathcal{L}_{reg}$ is changed to the L2 regular terms of the injected noise. Note that the detailed geometry can also be generated through the associations between geometry and texture established by $G_{detail}$. The latent code $z_{detail}$ mainly controls the generation of medium-grained face details like detailed shape of facial features, while the small-grained facial details like freckles is controlled by the injected noise. As shown in Fig.~\ref{fig:fit_pipeline}, the facial details can be generated after the optimization.

In the expression refinement phase, expression-related geometry changes will be further refined through the optimization of our pre-trained expression refinement generator, $G_{exp}$. Input with a conditional image composed of the expression offset UV map generated from the base model fitting phase and the output detailed geometry generated from the detail model fitting phase, the expression latent code $z_{exp}$ and the injected noise are also first randomly sampled and then optimized by differentiable rendering with the same loss terms with the detailed model fitting phase. After the final optimization, more detailed geometry like a smiling mouth are further refined, as presented in Fig.~\ref{fig:fit_pipeline}.

\section{Experiments}
\label{sec:exp}

\begin{figure}
    \centering
  	\includegraphics[width=0.95\linewidth]{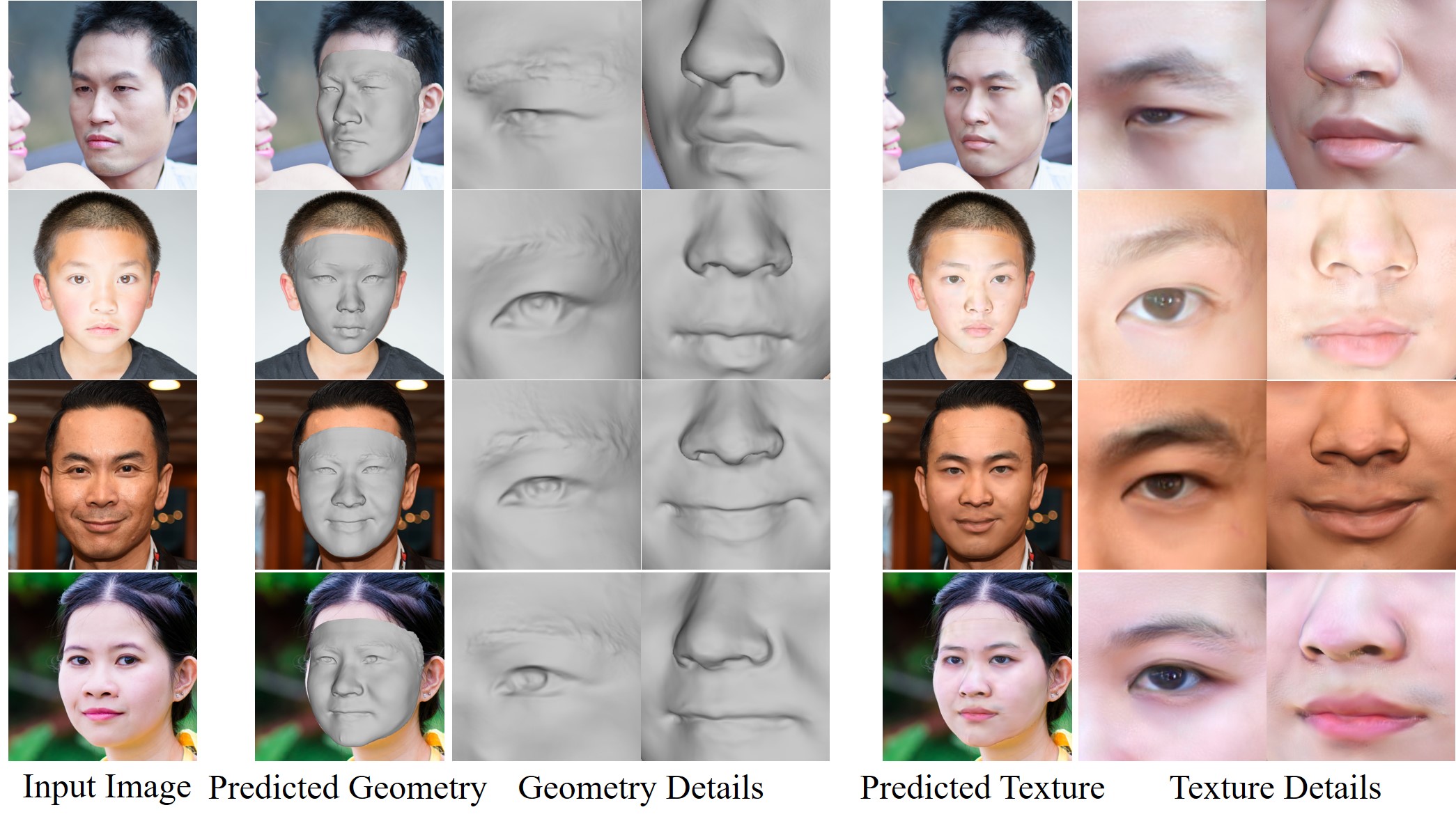}
    	\captionof{figure}{High-fidelity single-image fitting results predicted by our coarse-to-fine fitting pipeline.}
	 \label{fig:facepredict}
\end{figure}

\begin{figure}
    \centering 
  	\includegraphics[width=0.65\linewidth]{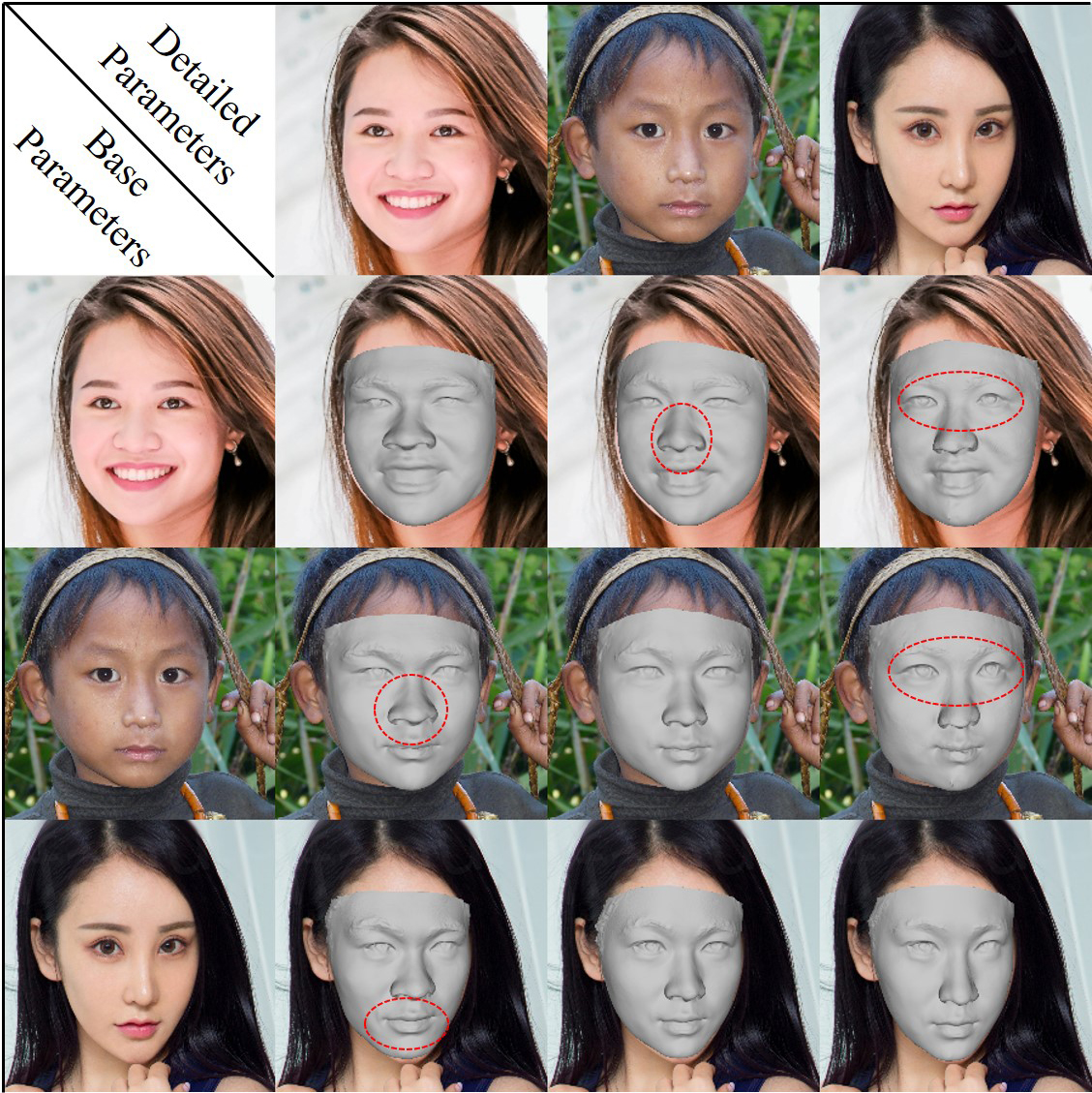}
    	\captionof{figure}{We use the base parameters fitted from the left images and detailed parameters fitted from the top images to generate new 3D face models, which keep the basic shape of the left faces and the detailed shape of facial features of the top images.}
	 \label{fig:detailtransfer}
\end{figure}%

\subsection{Evaluation}
\label{sec:evaluation}
We firstly evaluate the performance of our coarse-to-fine 3D model fitting framework in predicting 3D face model from a single face image. As shown in Fig.~\ref{fig:facepredict}, based on FaceVerse, our method shows both high-generalization and high-fidelity in predicting 3D face models from various input East Asian face images among different ages, genders or skin colors. On one hand, the base model built on large-scale base face dataset provide more prior knowledge in rough face fitting which makes the method more robust to various face images. On the other hand, the conditional styleGAN-based detailed model trained on the detailed dataset shows powerful ability in facial geometry and texture details generation. The texture details and geometry details in Fig.~\ref{fig:facepredict} proves that even the facial details in pupils and eyebrows can still be described by our details generator. 

In addition, both the base shape and the detailed shape of facial features can be adjusted by parameters in our method. To demonstrate the changeability of our model, we conduct a detail transfer experiment over the single-images fitting results, as shown in Fig.~\ref{fig:detailtransfer}. Using the base parameters fitted from images in the left column and the detail parameters fitted from images in the top row, our method can generate new face models which has the basic shape of the source face and the details of the target face (e.g. bigger eyes, thinner lips or a broader nose). Some images are sampled from the FFHQ dataset.

\subsection{Comparisons to Prior Works}
\label{sec:comparisons}

\begin{figure}
    \centering
  	\includegraphics[width=0.99\linewidth]{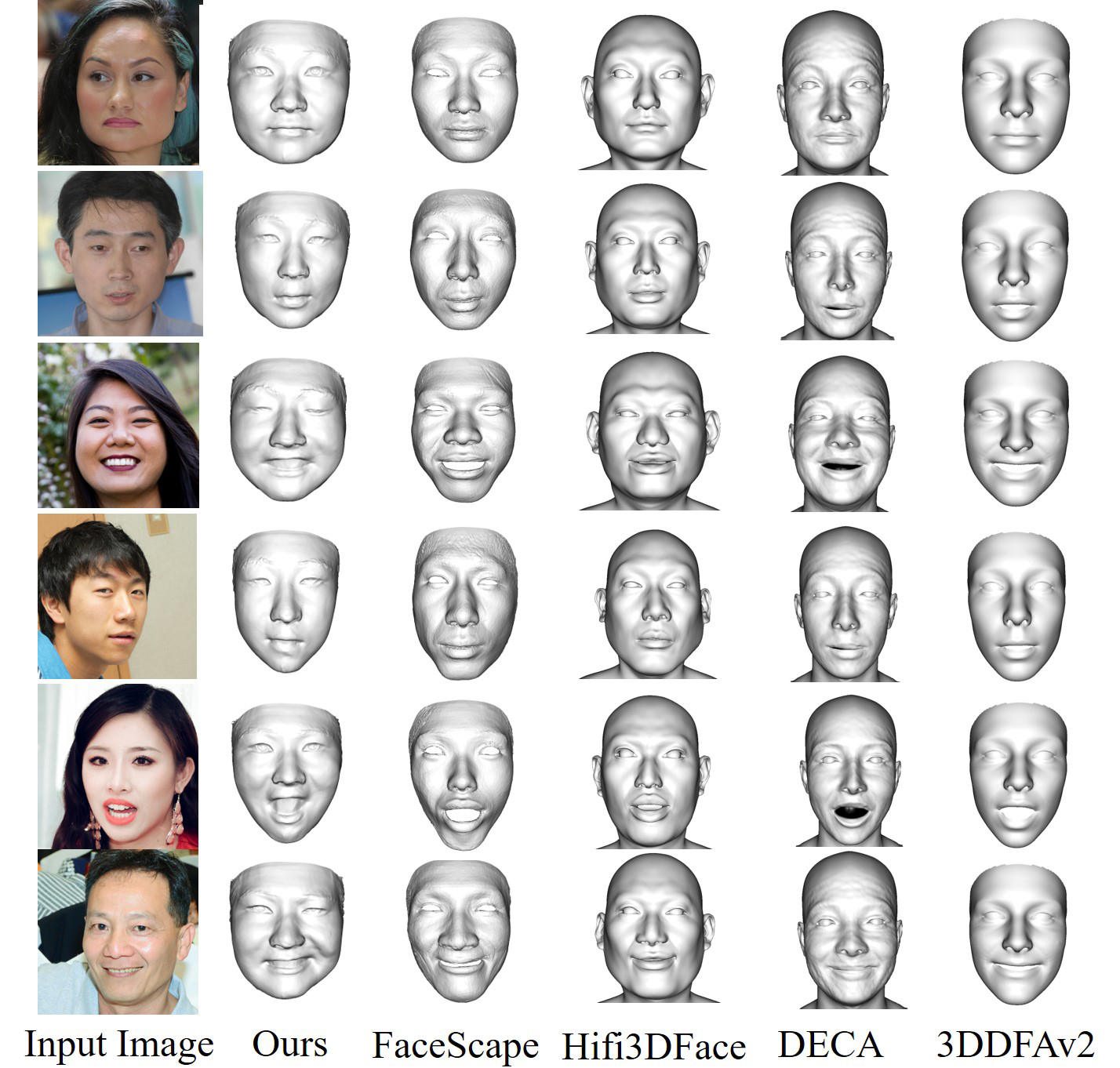}
    	\captionof{figure}{Comparison with the monocular 3D face reconstruction methods FaceScape, Hifi3DFace, DECAFace and 3DDFAv2.}
	 \label{fig:submainexp}
\end{figure}%

\begin{figure}
    \centering
  	\includegraphics[width=\linewidth]{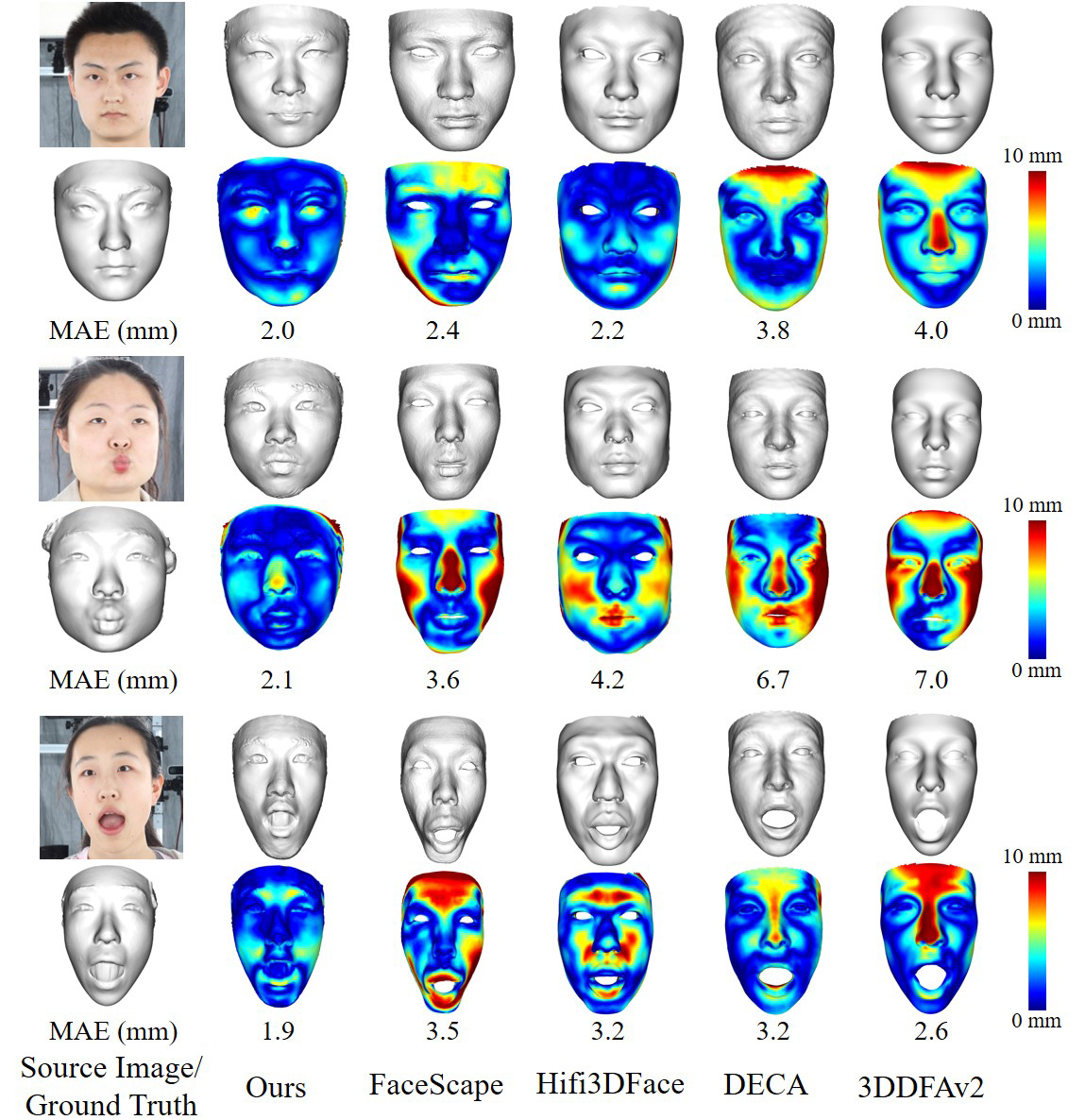}
    	\captionof{figure}{Quantitative Comparison of  3D face reconstruction methods. The length of ground-truth models is fixed at 200mm.}
	 \label{fig:icperror}
\end{figure}%

\begin{figure}
    \centering
  	\includegraphics[width=0.98\linewidth]{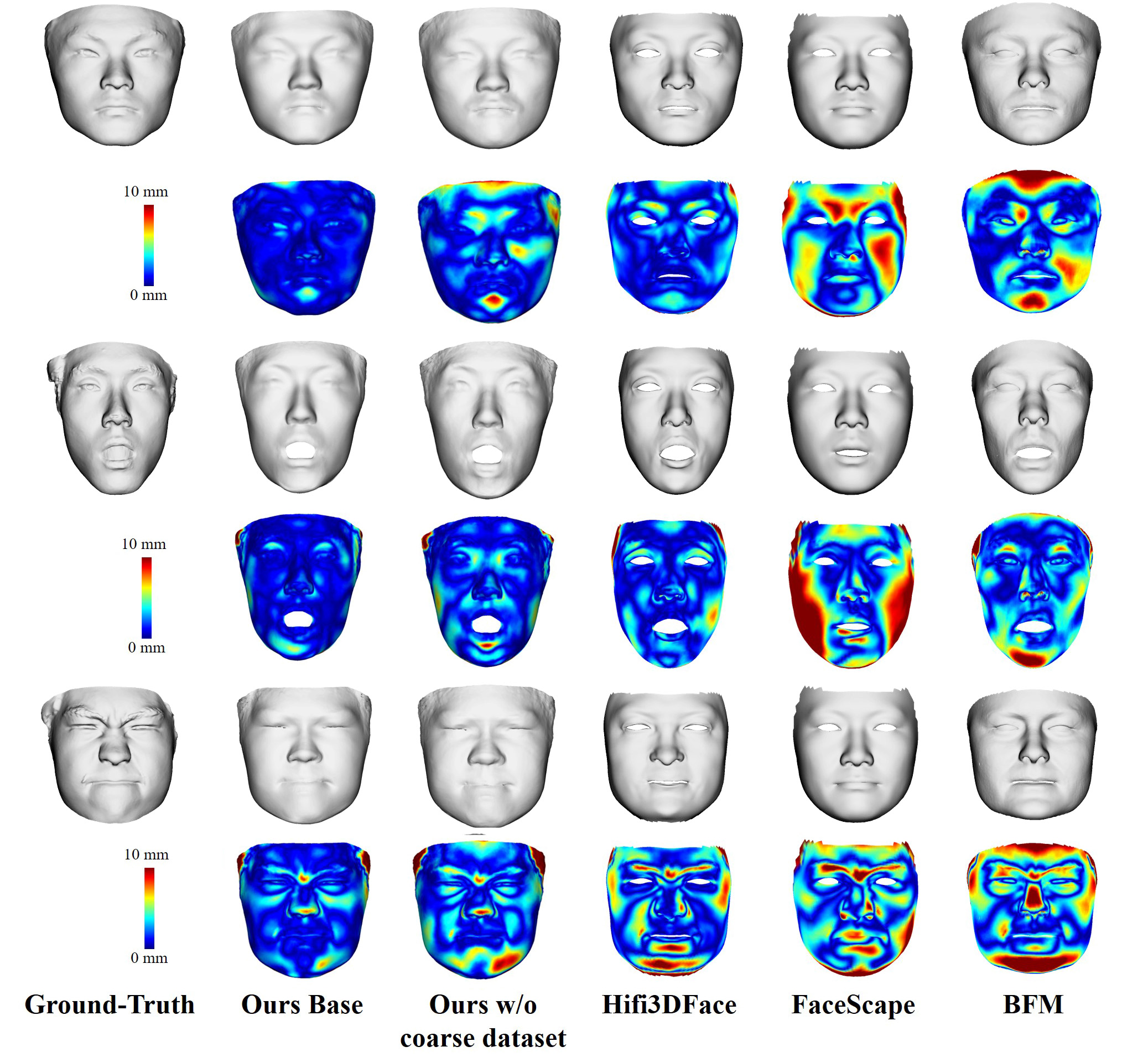}
    	\captionof{figure}{Quantitative comparison with 3DMM methods in 3D model fitting.}
	 \label{fig:exp}
\end{figure}
	
We compare our monocular fitting results with the state-of-art monocular facial reconstruction methods, including FaceScape~\cite{CVPR2020Facescape} and Hifi3DFace~\cite{bao2021tog} which are also proposed for East Asian facial reconstruction, as well as DECA~\cite{TOG2021deca} and 3DDFAv2~\cite{eccv2020towards} which is based on BFM~\cite{AVSBS2009BFM} and FLAME~\cite{TOG2017flame} respectively. As shown in Fig.~\ref{fig:submainexp}, gaining from the large-scale base model and GAN-based detail generator, our method shows better qualitative performance in both fitting face rough shape and generating face details compared with other methods. We also conduct a quantitative comparison using a single image and the corresponding detailed 3D models sampled from our testing set. As shown in Fig.~\ref{fig:icperror}, the generated models of different methods are fitted to the ground-truth models by a rigid-ICP algorithm. The calculated MAE error is displayed below the models and our method shows the best quantitative performance.

\begin{figure} 
\begin{minipage}{.54\linewidth} 
\footnotesize
	\centering
	\begin{tabular}{ccc}
	\toprule
		Method & MAE & Var\\
	\hline
		\textbf{Our base model} & \textbf{0.69} & \textbf{0.47} \\ 
		\textbf{Ours w/o coarse} & 1.08 & 1.26 \\ 
	\hline
		Hifi3DFace~\cite{bao2021tog}  &  1.08 & 1.38\\ 
		FaceScape~\cite{CVPR2020Facescape} &  1.74 & 2.60\\ 
		BFM~\cite{AVSBS2009BFM}  &  2.38 & 10.78\\ 

		\bottomrule
	\end{tabular}
\end{minipage}% 
\hspace{0.05\linewidth}
\begin{minipage}{.4\linewidth} 
\centering 
\includegraphics[width=\linewidth]{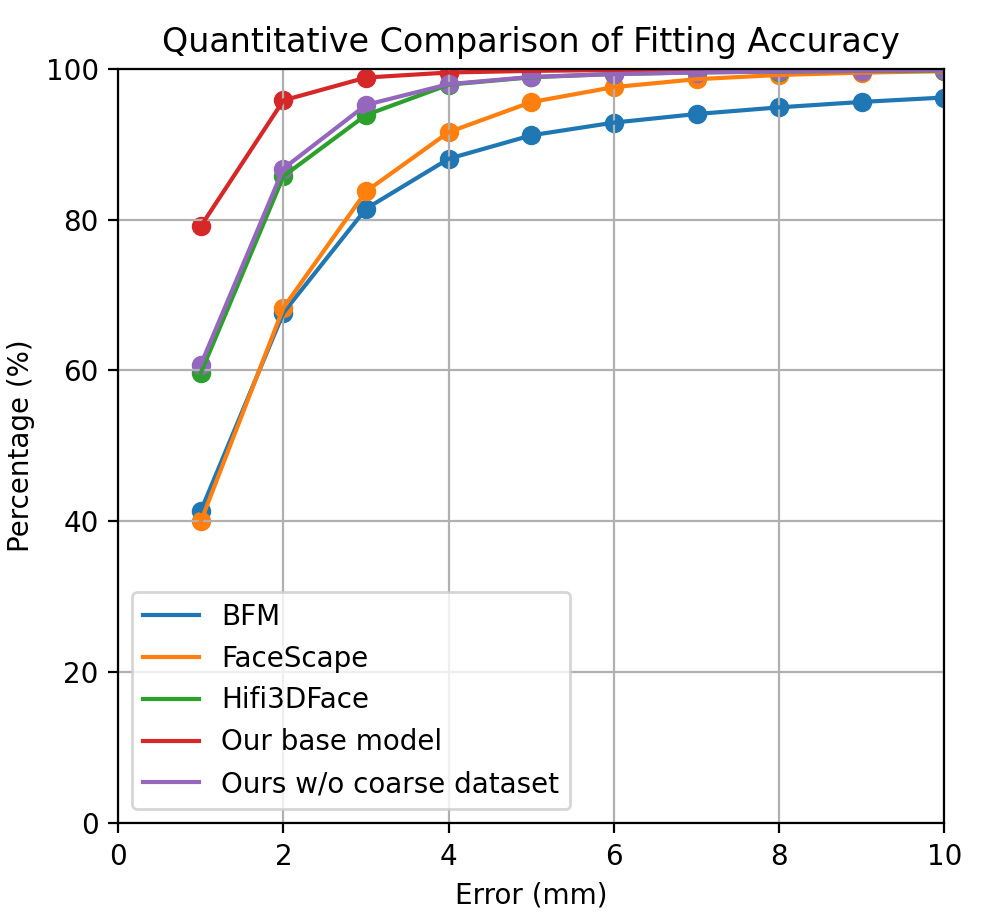} 
%\caption{Image} 
\end{minipage} 

\caption{Quantitative Comparison of our base model, FaceScape, Hifi3DFace and BFM in 3D model fitting. The left table presents the mean absolute error and variance in millimeter.}
\label{fig:icperrors}
\end{figure}

To further demonstrate the effectiveness of our parametric base model, we conduct a quantitative comparison with the state-of-the-art asian facial parametric models proposed by FaceScape~\cite{CVPR2020Facescape} and Hifi3DFace~\cite{bao2021tog}, as well as the BFM~\cite{AVSBS2009BFM}, on 3D scans from our testing set, which contains 357 models from 17 people and the models are fixed in the length of 200mm. We fit the parametric models to 3D scans by an optimization algorithm, which is based on the back-propagation through ICP (the algorithm is explained in detail in our supplementary pdf file). Note that our detailed model needs additional texture input, so we only use our base model to make a fair comparison. As shown in Fig.~\ref{fig:icperrors}, benefiting from the large-scale dataset, our base model shows the best quantitative performance in 3D model fitting. The visualized results are also presented in Fig.~\ref{fig:exp}.

\subsection{Ablation Study}
\label{sec:ablation}

\begin{figure}
    \centering 
  	\includegraphics[width=0.99\linewidth]{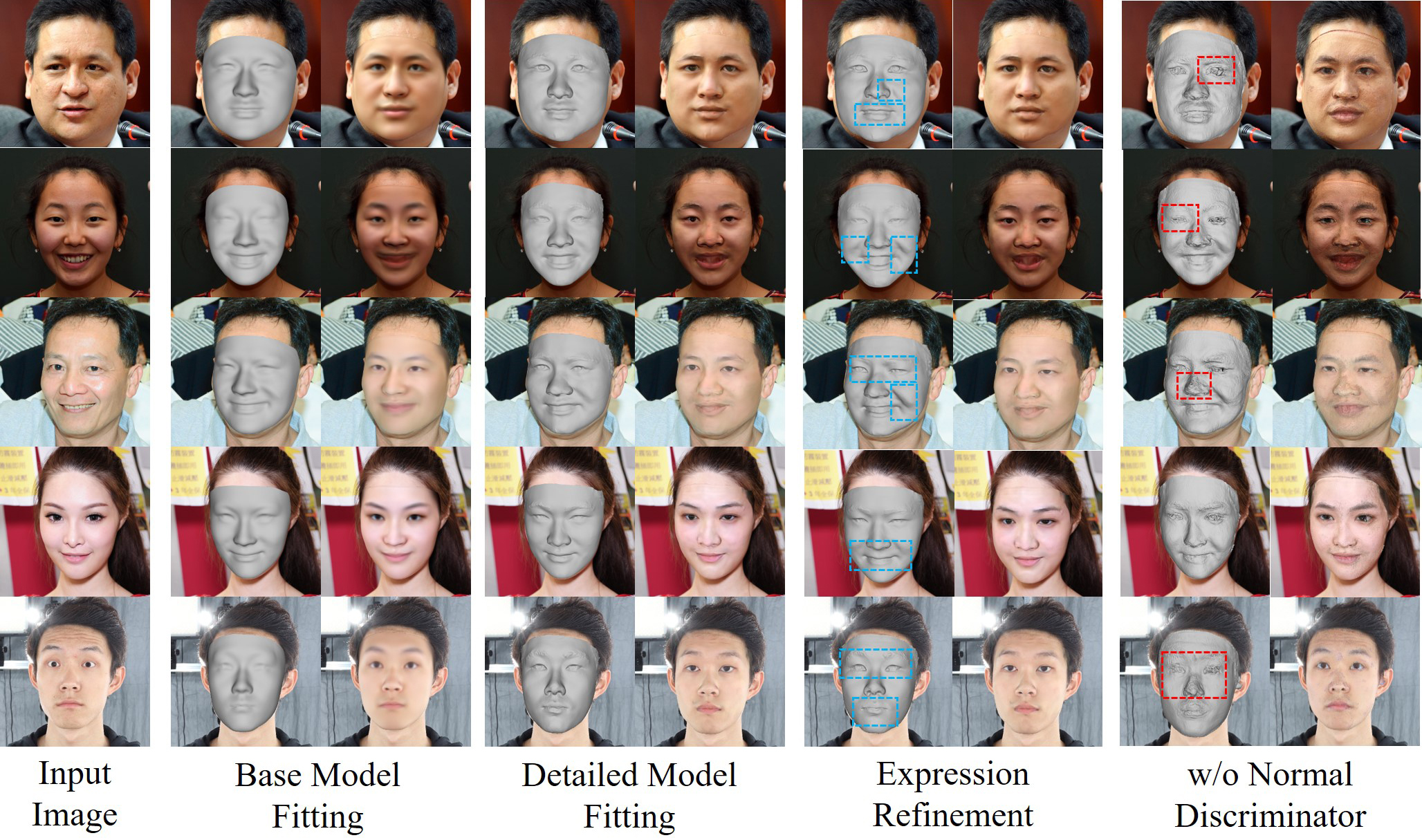}
    	\captionof{figure}{Monocular fitting results of our base model, detail generator, expression refinement generator and the model trained without the normal discriminator.}
	 \label{fig:ablationexp}
\end{figure}

In order to demonstrate the effectiveness of the modules used in our approach, we compare the fitting results of our base model, detailed results generated by the detail generator $G_{detail}$, refined results of the expression refinement generator $G_{exp}$ and results generated by the detailed model trained without our normal discriminator. As shown in Fig.~\ref{fig:ablationexp}, given a base shape and texture, our detail generator can add reasonable details but still lack description power of expressions. The geometric changes caused by expressions can be further refined by $G_{exp}$, as indicated by the blue rectangles in Fig.~\ref{fig:ablationexp}. Besides, the detailed model trained without our normal discriminator shows messy geometry, which demonstrates the effectiveness of our normal discriminator. Furthermore, the ablation study of the effects of the injected noise and latent code of our detailed model is presented in our supplementary materials. Please watch our supplementary video for more results. 

To further prove the superiority of introducing the coarse dataset into our base model, we generate an additional base model only using our detailed dataset, which contains 50 shape principal components and using the same expression principal components with our full base model. The 3D fitting results are also presented in Fig.~\ref{fig:exp} and Fig.~\ref{fig:icperrors} (labelled as ``Ours w/o coarse dataset''). The quantitative results prove that the fitting ability is significantly improved after introducing our coarse dataset.

\section{Discussion and Conclusion}
\label{sec:conclusion}

\begin{figure}
    \centering 
  	\includegraphics[width=0.99\linewidth]{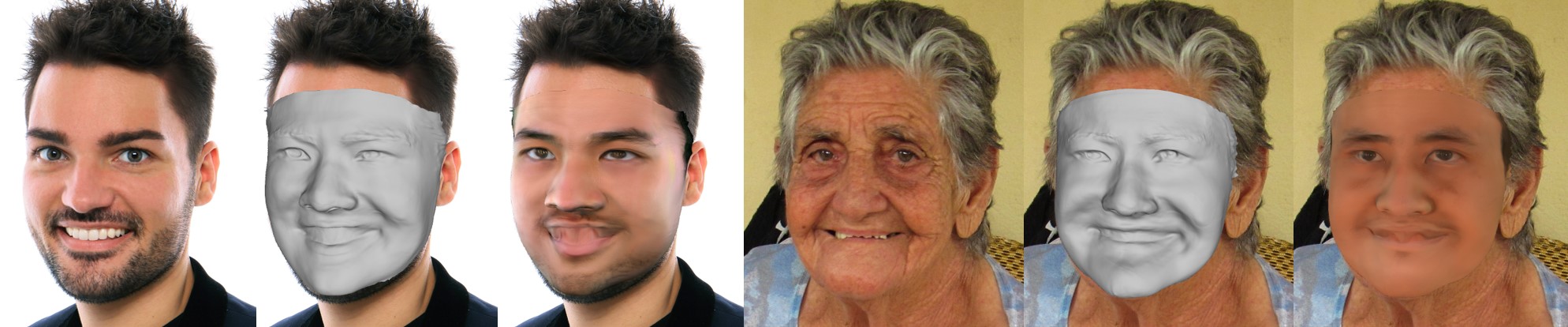}
    	\captionof{figure}{Limitations of our method. The proposed FaceVerse can not generate a thick beard and deep wrinkles.}
	 \label{fig:limit}
\end{figure}

\noindent\textbf{Limitations.}  On one hand, our dataset only contains faces of East Asians, and thus our performance declines when fitting the faces from other regions. On the other hand, our detailed model still suffers from a lack of detailed 3D face scans of the old people. As a result, as shown in Fig.~\ref{fig:limit}, our method can not generate extreme textures like a thick beard and can not generate deep wrinkles of the old people.

\noindent\textbf{Potential Social Impact.}  Our method enables 3D face reconstruction from a single image. Therefore, it can be used to generate a 3D fake model of a person, which needs to be addressed carefully before deploying the technology.

\noindent\textbf{Conclusion.}  In this paper, we have presented FaceVerse, a fine-grained and detail-changeable 3D face morphable model from a hybrid dataset. We have collected a large-scale coarse dataset and a high-fidelity detailed dataset and proposed a coarse-to-fine scheme to build our model, which guarantees the high generalization ablity and high fidelity of our model. The proposed conditional StyleGAN is able to generate and control the facial geometry and texture details while perserving the basic facial atrributes from the base model. Experiments have demonstated the superiority of our method in 3D face model fitting and monocular face reconstruction compared with the state-of-the-art methods. We believe FaceVerse can be a powerful tool for face-related researches and our pipeline will inspire the following research of 3DMM and monocular 3D facial reconstruction.

\noindent\textbf{Acknowledgement.}  This work is supported by Ant Group through Ant Research Program and is sponsed by NSFC No. 62125107 and No. 62171255.

%%%%%%%%% REFERENCES

\end{document}